\PassOptionsToPackage{expansion=false}{microtype}
\documentclass[]{pkuai4m}

\usepackage[toc,page,header]{appendix}

\usepackage{multirow}
\usepackage{multicol}
\usepackage{booktabs}
\usepackage{colortbl}
\usepackage[table]{xcolor}
\usepackage{makecell}
\usepackage{graphicx}
\usepackage{xcolor}
\usepackage{fontawesome5}
\usepackage{caption}
\usepackage{fancyvrb}
\usepackage{fvextra}
\usepackage{enumitem}
\usepackage{wrapfig}
\usepackage{mathtools}
\usepackage{needspace}

\usepackage[most]{tcolorbox}
\usepackage{amsthm,amsmath,amssymb}

\theoremstyle{plain}
\newtheorem{theorem}{Theorem}

\newtheorem*{theorem*}{Theorem}
\newtheorem*{problem*}{Problem}
\newtheorem*{remark*}{Remark}
\newtheorem*{claim*}{Claim}
\newtheorem*{conjecture*}{Conjecture}

\theoremstyle{definition}

\tcolorboxenvironment{theorem}{
    colback=blue!5,
    colframe=blue!5,
    arc=1mm,
    fonttitle=\bfseries,
    boxrule=0pt,
}
\tcolorboxenvironment{proposition}{
    colback=blue!5,
    colframe=blue!5,
    arc=1mm,
    fonttitle=\bfseries,
    boxrule=0pt,
}
\tcolorboxenvironment{lemma}{
    colback=blue!5,
    colframe=blue!5,
    arc=1mm,
    fonttitle=\bfseries,
    boxrule=0pt,
}
\tcolorboxenvironment{corollary}{
    colback=blue!5,
    colframe=blue!5,
    arc=1mm,
    fonttitle=\bfseries,
    boxrule=0pt,
}
\tcolorboxenvironment{definition}{
    colback=red!5,
    colframe=red!5,
    arc=1mm,
    fonttitle=\bfseries,
    boxrule=0pt,
}
\tcolorboxenvironment{problem*}{
    colback=orange!5,
    colframe=orange!5,
    arc=1mm,
    fonttitle=\bfseries,
    boxrule=0pt,
}
\tcolorboxenvironment{conjecture*}{
    colback=orange!5,
    colframe=orange!5,
    arc=1mm,
    fonttitle=\bfseries,
    boxrule=0pt,
}
\tcolorboxenvironment{cited}{
    colback=green!5,
    colframe=green!5,
    arc=1mm,
    fonttitle=\bfseries,
    boxrule=0pt,
}

\title{Danus: Orchestrating Mathematical Reasoning Agents with Fact-Graph Memory}

\author[1,2,*]{Jihao Liu}
\author[1,*]{Guoxiong Gao}
\author[2,3,*]{Zeming Sun}
\author[4,5,*]{Bin Wu}
\author[6]{Shurui Liu}
\author[7]{Jiedong Jiang}
\author[1]{Haocheng Ju}
\author[1]{Leheng Chen}
\author[6]{Ronnie Cheng}
\author[8]{Xiping Zhang}
\author[5,9,10,11\P]{Bin Dong}

\affiliation[]{
$^{1}$School of Mathematical Sciences, Peking University \\
$^{2}$Beijing International Center for Mathematical Research,  Peking University \\
$^{3}$Research Institute for Mathematical Sciences, Kyoto University\\
$^{4}$School of Mathematics, Tianjin University \\
$^{5}$Zhongguancun Academy \\
$^{6}$Department of Mathematics, Stanford University \\
$^{7}$Westlake Institute for Advanced Study, Westlake University  \\
$^{8}$School of Mathematical Sciences, Key Laboratory of Intelligent Computing and Applications (Ministry of Education),  Tongji University\\
$^{9}$Beijing International Center for Mathematical Research and the New Cornerstone Science Laboratory, Peking University \\
$^{10}$Center for Machine Learning Research, Peking University\\
$^{11}$Center for Intelligent Computing, Great Bay Institute for Advanced Study, Great Bay University \\
}

\contribution[*]{Equal Contribution}
\contribution[\P]{Corresponding author}

\DeclareFontShape{T1}{bytesans}{b}{n}{<-> s * [1] pkuai4m/bytesans}{}
\DeclareFontShape{T1}{bytesans}{bx}{n}{<-> s * [1] pkuai4m/bytesans}{}
\abstract{
Recent LLM-based mathematical reasoning agents have begun to tackle research-level problems and, in several cases, have contributed to the resolution of open problems. However, scaling and orchestrating such agents effectively remains challenging, due to the difficulty of coordinating parallel proof search while keeping intermediate claims organized and reliable. In this paper, we propose Danus, an orchestration system for research-level mathematical reasoning centered on a shared fact graph as a global memory-management mechanism. Danus consists of a main agent that performs planning and coordination, multiple worker agents that carry out proof search in parallel, and a stateless verifier that checks proposed mathematical claims before they are admitted into the fact graph. Each verified fact is stored together with its proof and logical dependencies, allowing the system to build long arguments incrementally while keeping the shared proof state organized. The main agent periodically summarizes the evolving proof state, redirects workers across promising directions, and supports interaction with human mathematicians through progress reports. We evaluate Danus through six research-level case studies in algebraic geometry, singularity theory, and combinatorics, illustrating how the fact-graph memory mechanism enables Danus to construct long, detailed mathematical proofs. Our results suggest that fact-graph-based orchestration provides an effective route toward scaling mathematical reasoning agents for long-horizon research problems. Danus is open source.

}

\date{\today}

\def\emailicon{\raisebox{-1.5pt}{\includegraphics[height=1.05em]{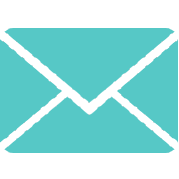}}}
\def\githubicon{\raisebox{-1.5pt}{\includegraphics[height=1.05em]{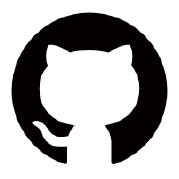}}}

\checkdata[ \emailicon \hspace{0.3em} Correspondence ]{\email{dongbin@math.pku.edu.cn}}

\newcommand{\danuslink}{https://github.com/frenzymath/Danus}

  \checkdata[ \githubicon \hspace{0.3em} Danus Source]{ \url{\danuslink} }

\begin{document}
\renewcommand{\today}{July 7, 2026}
\maketitle

\renewcommand{\thefootnote}{\fnsymbol{footnote}} 
\setcounter{footnote}{0}

\renewcommand{\thefootnote}{\arabic{footnote}}
\pagestyle{fancy}
\fancyhf{}
\fancyhead[R]{\thepage}

\section{Introduction}

Large language models (LLMs) are increasingly deployed as the reasoning cores of agentic systems that retrieve knowledge, call tools, execute code, interact with external environments, and revise their outputs through feedback. In these systems, performance depends not only on the base model, but also on the harness that governs how information, tools, state, and feedback are exposed to the model. With appropriate harness engineering, LLM-based agents can outperform non-interactive prompting baselines on tasks requiring external information, iterative correction, or long-horizon execution \cite{lee2026metaharnessendtoendoptimizationmodel,yang2024sweagent}. 

Several agents have also been proposed for research-level mathematical reasoning. The Aletheia agent \cite{feng2026towards}, built on an advanced version of Gemini Deep Think, consists of a generator, verifier, and reviser, and iterates over these three components. It has resolved several Erdős problems either autonomously or semi-autonomously, and has also been applied to research-level problems in algebraic geometry \cite{patel2026simplicity}, combinatorics \cite{lee2026lower}, and representation theory \cite{feng2026eigenweights}. The Rethlas agent \cite{Ju+26} is designed to mimic the workflow of human mathematicians. It is equipped with skills and tools tailored to mathematical research, and maintains a working memory of intermediate artifacts generated during the reasoning process, such as constructed examples, counterexamples, and subgoal-decomposition plans. Rethlas has autonomously addressed several open problems in commutative algebra \cite{jiang2026openproblemscommutativealgebra}, functional analysis \cite{dou2026degenerateconstantsdegreeinequalities}, and probability \cite{li2026injectivity}, and has also assisted in resolving mathematical research problems in algebraic geometry \cite{pan2026lift}. The QED agent \cite{an2026qedopensourcemultiagentgenerating} consists of a decomposer, prover, structural verifier, detailed verifier, and regulator, and orchestrates these purpose-specific agents in loops. It has resolved several research-level problems in algebraic geometry, PDEs, probability, and inverse problems. The ProofCouncil system \cite{schmitt2026proofcouncil} consists of an author agent, an advisory LLM council, a critic agent, and a compute agent, and iterates over these components. It correctly resolved 6 out of 10 problems in the second batch of FirstProof, up to at most minor revisions \cite{abouzaid2026first}. The AI co-mathematician \cite{zheng2026ai} is an agentic AI workspace that coordinates multiple specialized agents to decompose problems, explore ideas, search the literature, run computations, draft informal proofs, and iteratively review and revise their outputs. It has helped human mathematicians resolve open problems in group theory in an interactive manner.

Most of the agents mentioned above incorporate, explicitly or implicitly, a generate--verify--revise loop, and the ``multi'' in existing multi-agent mathematical reasoning systems typically refers to agents with different specialized roles. By contrast, there has been little systematic study of how to scale mathematical reasoning systems built around generate--verify--revise loops by increasing the number of agents directly involved in proof generation. Scaling such agents is not simply a matter of spawning many agents to work on the same problem without further modification. It requires careful memory management and coordination. If the shared memory of multiple agents is not handled appropriately, it can confuse the agents, propagate irrelevant or erroneous intermediate artifacts, and ultimately hurt performance.

To scale and orchestrate mathematical reasoning agents effectively, we propose Danus, an orchestration system centered on a shared fact graph that serves as a global memory-management mechanism. In Danus, a main agent performs planning and coordination, while multiple worker agents carry out proof generation. The shared fact graph keeps intermediate informally verified facts organized, enabling the system to construct long, elaborated proofs. 

The interaction between Danus and human mathematicians is also carefully designed. The main agent regularly summarizes the current proof state into a report, allowing human mathematicians to inspect the progress of the proof and provide high-level guidance through the main agent when necessary. The same report can also be used by the main agent to consult advanced systems such as GPT-5.5-pro for additional mathematical guidance. In addition, Danus includes a writing system that turns the completed proof into a paper-style exposition once the target statement has been proved, making the resulting argument more accessible to human readers. We demonstrate the effectiveness of Danus on several challenging research-level problems in algebraic geometry, singularity theory, and combinatorics, and discuss the role of the fact-graph memory mechanism in producing long, detailed, and correct proofs. We also explicitly report the human input provided for each problem and how the mathematicians collaborated with Danus in these examples.

The remainder of the paper is organized as follows. Section \ref{sec:method} presents the design of the Danus system; Section \ref{sec:results}, six case studies; Section \ref{sec:discussion}, a discussion of its strengths and limitations; and Section \ref{sec:concl} concludes the paper.

\section{Methodology}\label{sec:method}

\begin{figure}[htb]
\centering
\includegraphics[width=\linewidth]{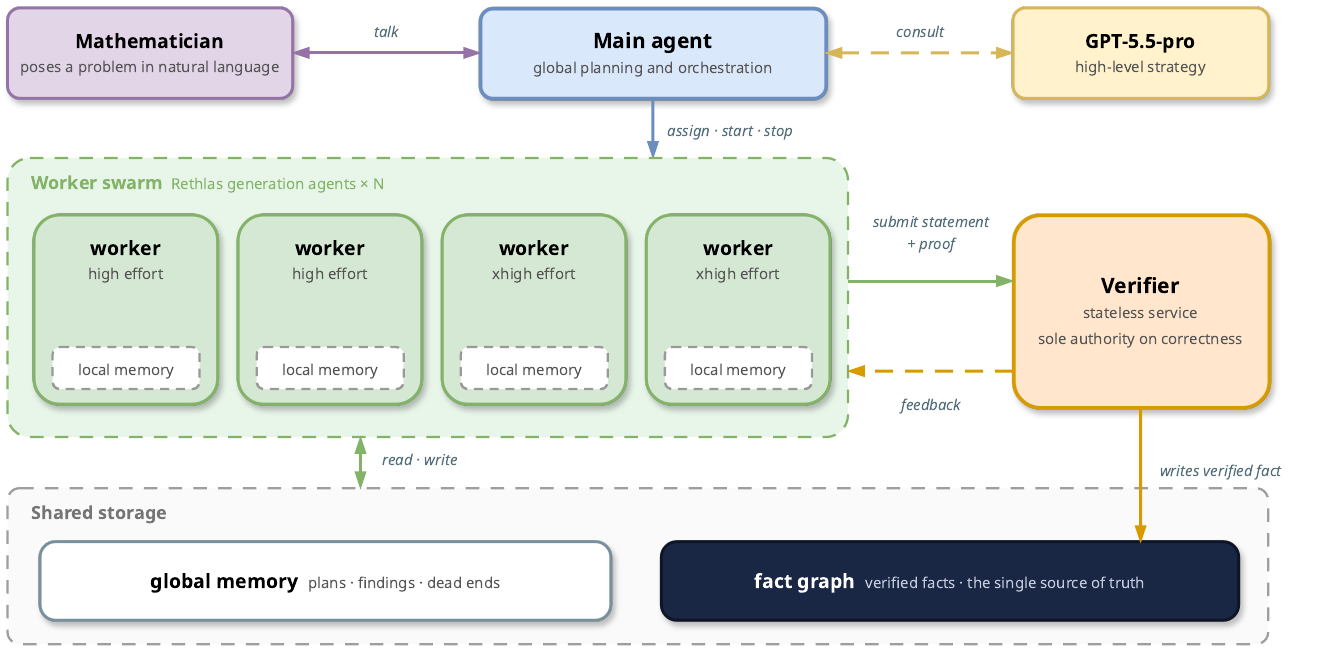}
\caption{The overall architecture of Danus.}
\label{fig:architecture}
\end{figure}

Danus is an automated system for research-level mathematics, built on the worker--verifier core of the previous system, Rethlas~\cite{Ju+26}. It couples a swarm of proof-search workers and a stateless verifier with a shared memory, all coordinated by a main agent. The design follows a strict separation of powers: the main agent performs the global planning and coordination, the workers carry out the detailed proof search, the verifier is the sole authority on correctness, and a single fact graph holds every verified result and is the system's only source of truth (Figure~\ref{fig:architecture}). The remainder of this section takes up each of these in turn: Section~\ref{sec:workflow} describes the workflow; Sections~\ref{sec:factgraph} and~\ref{sec:memory}, the fact graph and the memory beyond it; Sections~\ref{sec:mainagent} and~\ref{sec:workers}, the main agent, the workers, and the verifier; Section~\ref{sec:tools}, the skills and tools of each kind of agent; and Section~\ref{sec:writing}, summarization and paper writing.

\subsection{Workflow overview}\label{sec:workflow}

These parts work together on a single problem, from the initial statement to a finished paper. A mathematician gives the main agent a problem in natural language. The main agent forms an initial plan, consulting GPT-5.5-pro for high-level mathematical strategy when needed, and assigns directions to the workers. The workers are Rethlas generation agents that run in parallel and explore the problem from several directions, including both constructive and refutational routes. Each worker repeatedly proposes a verifiable statement together with a supporting proof and submits it to the verifier; a statement whose proof passes verification is stored as a \textit{fact}. These facts form a directed acyclic graph (DAG), called the \textit{fact graph}, whose edges record logical dependencies. As the workers proceed, the main agent periodically reads their progress and the fact graph, summarizes the current state, consults GPT-5.5-pro, and re-assigns the workers. The iteration is not stopped after a preset number of rounds; it stops only when the main agent confirms that the target statement, or its refutation, has appeared as a verified fact in the fact graph. The main agent can then turn the fact graph into a paper for human experts to check.

This parallel exploration is the main departure from Rethlas. Where Rethlas pursues one line of reasoning at a time, Danus runs several workers at once, each exploring a different facet of the problem, such as a different lemma to prove, a counterexample to construct, or a toy example to study. This broadens the exploration and makes long, multi-step arguments tractable. It also raises the problem that the fact graph is designed to solve: letting many workers contribute to a single proof without interfering with one another.

\needspace{0.45\textheight}
\subsection{The fact graph}\label{sec:factgraph}

\setlength{\intextsep}{0pt}%
\setlength{\columnsep}{8pt}%
\begin{wrapfigure}{l}{0.35\linewidth}
\centering
\captionsetup{skip=3pt}%
\includegraphics[width=\linewidth, trim=0 0 0 62, clip]{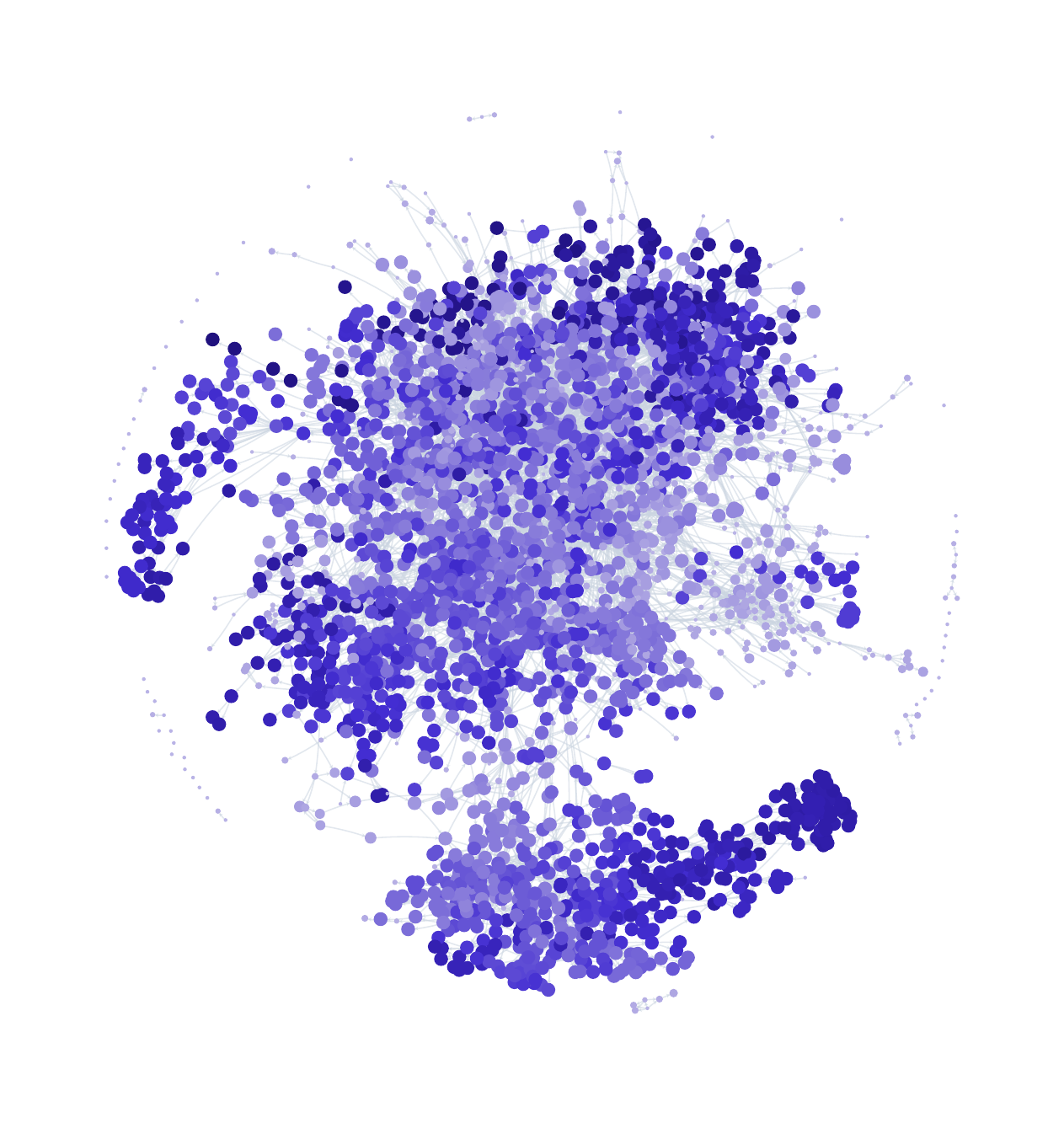}
\caption{The fact graph behind the matroid tangent-class case study of Section~\ref{sec:res-tangent}: 3,157 verified facts, 8,616 dependency edges; nodes darken and grow with dependency depth (up to 54). The clusters are separate lines of attack: bottom, conditional scaffolding that the final proof never cites; left, an independent re-derivation of the Chern-number bound; upper right, the integral lift of the result, whose final route enters the proof.}
\label{fig:factgraph}
\end{wrapfigure}%
\setlength{\intextsep}{12pt plus 2pt minus 2pt}%
\setlength{\columnsep}{1.2em}%
\textbf{Structure.} The fact graph is the unique source of truth of the whole system and its central design element. It is a directed acyclic graph (DAG) whose nodes are facts and whose edges record logical dependencies. A fact is a mathematical statement together with a verifier-checked proof; an edge from one fact to another means that the proof of the second uses the first. A worker can draw on the facts already in the graph: in principle it has access to the entire graph, and in practice it retrieves the relevant facts by searching it. When a worker submits a new statement and proof, it records the identifiers of the facts that the proof depends on, and these become the new fact's incoming edges. Every statement that passes verification is added to the graph, which therefore grows until it contains the target theorem (Figure~\ref{fig:factgraph}).

\textbf{Why a graph rather than a single blueprint.} This design is what lets many workers collaborate. In Rethlas, the whole result was carried by a single Markdown blueprint that held all supporting lemmas, definitions, and the final theorem; a worker repeatedly edited this blueprint and asked the verifier to check it and suggest revisions. Compared with a single blueprint, the fact graph has two advantages. First, it keeps each worker's context small and focused. Such context management matters because a language-model agent reasons most reliably over a short context holding only relevant material; unrelated content both strains its capacity and interferes with its reasoning. A single blueprint forces every worker to carry the whole accumulated proof, whereas the fact graph lets each worker draw on only the facts it needs for its current claim and submit one fact at a time, so the working context stays small even as the proof grows to many pages. Second, it supports parallel work: a single file is hard for many workers to edit at once or to use for separate lines of attack, while the fact graph lets their contributions accumulate into one shared structure.

\textbf{Revocation.} The fact graph also supports revocation: a fact that is later found to be wrong is removed together with every fact that depends on it, directly or transitively. This happens in one of two situations: an error in a cited reference (usually found by the agent during a final review, or pointed out by a human expert) or, more rarely, a conceptual confusion or a flawed proof. In our runs, revocation was needed only very rarely, a point we return to when we discuss the reliability of verification.

\subsection{Memory beyond the fact graph}\label{sec:memory}

Content that has been verified lives in the fact graph; everything else worth keeping lives in memory. Memory is not part of the truth; it is shared context that helps the workers avoid repeating one another's failed attempts, and it has two tiers. Each worker keeps a private local memory, a running log of its own activity, used mainly for later analysis of how a line of reasoning developed. Above it is a global memory that every worker and the main agent can read and write. The global memory records the intermediate products of the search, such as plans, promising directions, dead ends, and constructed examples and counterexamples. It also records the prompt and reply of every GPT-5.5-pro consultation. Together, the two tiers keep each worker from repeating its own local exploration and let the main agent see what has happened along each worker's branch of the search, so that it retains an overall grasp of the reasoning strategy and the plan.

\subsection{The main agent}\label{sec:mainagent}

\textbf{Role.} The main agent, the workers, and the verifier are the three kinds of agent that act on the fact graph and the memory; of these, the main agent is the principal addition relative to Rethlas. It is responsible for the global planning and orchestration of the search: it forms and continually revises the overall proof strategy, decides how to decompose the problem and allocate effort across the workers, judges when a line of attack should be abandoned in favor of a new one, and conducts the strategic interaction with the human expert; it also performs delegated tasks such as writing up the results. Its work is confined to this global level: the main agent does not itself carry out the detailed mathematical derivations, which are left to the workers and constitute the only content subject to verification. The system thereby maintains a separation of powers, preventing the agent that steers the search from introducing unverified mathematics into the fact graph. The main agent forms these judgments from its own understanding of the overall structure, its observation of each worker's progress, and the contents of the global fact graph and memory. For fine points of mathematics it may consult GPT-5.5-pro at low frequency, at most once per hour, as an expert reference. Its strategic decisions therefore rest on a system-wide view of the run, augmented where needed by the mathematical strength of a frontier model.

\textbf{The strategy loop.} In operation, the main agent first interprets the task set by the human expert, consults GPT-5.5-pro for high-level strategic guidance, and assigns initial directions to the workers. At regular intervals (every one to two hours), it re-reads the workers' logs, the memory, and the fact graph to understand the current state and the points at which the search is stuck, produces a summary of this state, and asks GPT-5.5-pro for the next strategic step. This cycle repeats, and when the main agent confirms that the problem has been solved it notifies the human expert and supports downstream tasks such as paper generation. Throughout, the human can interact with the main agent at any time, to ask about progress, redirect the search, or adjust priorities, without interrupting the running workers. Section~\ref{sec:writing} describes how the progress and the final result are presented to the human.

\textbf{Reading the global state.} A hard part of this role is the periodic summary. The main agent must read through many layered log files and a fact graph that can hold thousands of entries, follow the progress of several workers at once, and combine all of it into a single summary that a human or GPT-5.5-pro can act on. This reading over a large, evolving body of files calls for the capabilities of a coding agent, which is built to navigate and read extensive bodies of code. We evaluated several such agents (Codex, Claude Code, and OpenClaw, loaded with GPT-5.5 or Claude Opus 4.8) and selected Claude Code with Claude Opus 4.8, which handled this reading task best; the main agent is built on it.

\subsection{Workers and the verification service}\label{sec:workers}

\textbf{Workers.} While the main agent is new, the worker and verifier are inherited, with only minor changes to the agents themselves, from the generation and verification agents of Rethlas; what changes is chiefly how the worker is used. A worker now works under a task assigned by the main agent and typically focuses on one claim at a time, such as a lemma, a counterexample, or a toy example, rather than an entire proof. It repeatedly submits this claim to the verifier and revises it under the verifier's feedback until it passes, at which point the claim enters the fact graph as a fact. Because the worker retains Rethlas's core retrieval mechanism, built on the theorem search engine Matlas~\cite{Matlas26}, it can search the mathematical literature precisely for relevant results. In practice we run three to nine workers per project and split them roughly evenly between two reasoning-effort levels, ``high'' and ``xhigh'' (the effort setting of the underlying Codex agent). Reserving half of the workers for the lower ``high'' effort is deliberate: it supplements the search with shallower but useful conclusions and increases the diversity of reasoning across the swarm.

\begin{figure}[htb]
\centering
\includegraphics[width=\linewidth]{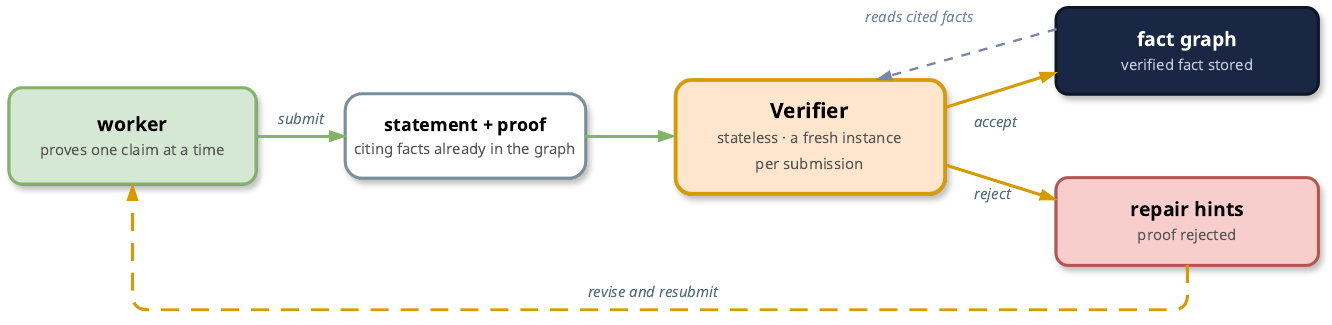}
\caption{The submit--verify--repair cycle between a worker and the verifier.}
\label{fig:verify-loop}
\end{figure}

\textbf{The verifier.} The verifier is run as a service that both the workers and the main agent can call, and it is stateless: a fresh instance judges each submission and retains nothing afterwards. Crucially, it is permitted to read the fact graph, so that when a proof depends on other facts it can follow those citations and check the dependency. Owing to the skills and checking procedure carefully engineered in Rethlas, the verifier produced essentially no false positives on the problems we tested; in the rare cases where it did, it accepted a proof containing a small number of skipped steps, or, because it takes cited references to be correct, a proof that relied on an erroneous reference. Both kinds of slip are easy to catch in a final review, which is what makes the fact graph reliable as the system's source of truth. The full submit--verify--repair cycle is shown in Figure~\ref{fig:verify-loop}.

\subsection{Skills, tools, and interfaces}\label{sec:tools}

\begin{figure}[htb]
\centering
\includegraphics[width=\linewidth]{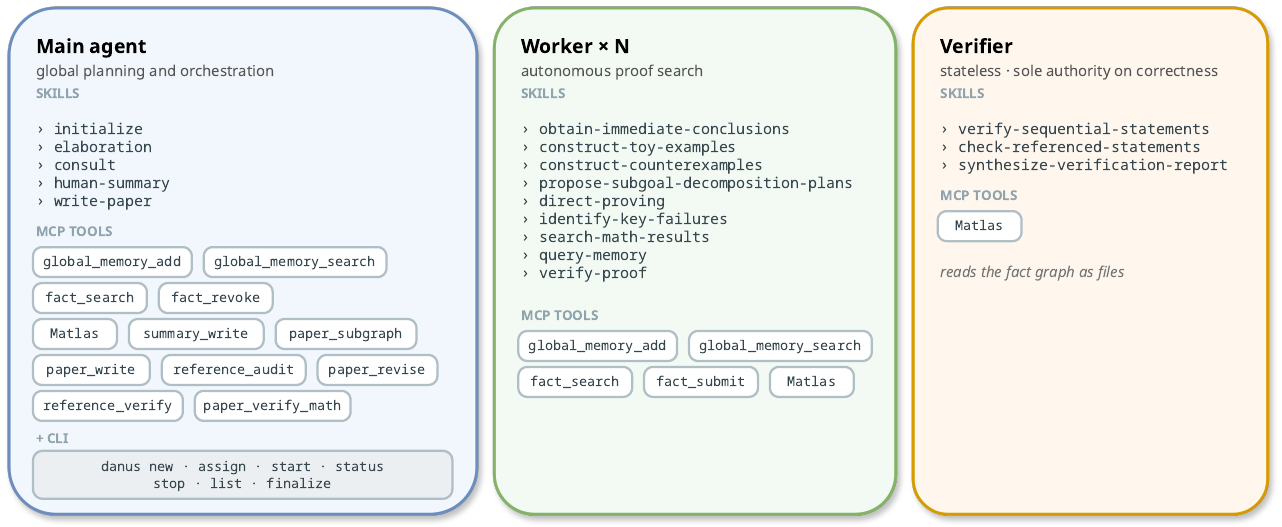}
\caption{The three kinds of agent, each with its own skills and its own role-gated set of tools.}
\label{fig:agent-tools}
\end{figure}

Each kind of agent carries out its role through its own skills and its own tools (Figure~\ref{fig:agent-tools}). The worker and verifier skills are inherited essentially unchanged from Rethlas: the worker's are reasoning primitives such as constructing examples and counterexamples, decomposing a goal into subgoals, and direct proving, while the verifier's are proof-checking primitives such as checking a proof step by step and confirming that cited statements exist and are applicable. The main agent has its own skills for consulting GPT-5.5-pro, producing summaries, and writing. A small command-line interface lets the main agent start, assign work to, monitor, and stop the workers, and a set of tools exposed through the Model Context Protocol (MCP) lets the agents interact with the fact graph and the global memory. Retrieval from the mathematical literature is provided as one such tool, backed by Matlas, and is available to the workers, the main agent, and the verifier alike.

\subsection{Summarization and paper writing}\label{sec:writing}

Two further components produce the system's output for the human experts: summarization while a run is in progress, and paper writing once the proof is complete. The two differ in purpose. Summarization presents the unfinished search through a progress report, so that the experts can follow and steer it; paper writing turns the verified fact graph into a readable manuscript rather than a list of facts, so that they can check the result quickly.

\textbf{Summarization.} The progress report is written by an isolated agent that receives only the problem statement and the verified mathematics, stripped of every reference to the system's internals, so the report cannot mention what its author never saw. The skill that governs it is a set of prohibitions rather than a guide to fluent prose: progress is never estimated numerically, and a result is reported as proven only when a verified fact settles it with no hypothesis left to match; the default is to report it as weaker. The prohibitions exist because a model left to itself summarizes optimistically---a conditional argument with one unmatched hypothesis becomes ``essentially complete''---and the report is the experts' window into the run. The same discipline, with one further rule, governs the state summaries that the main agent writes for GPT-5.5-pro (Section~\ref{sec:mainagent}): there, every stalled line must be classified as a failure of the method, which leaves the statement alive, or as evidence against the statement itself, the distinction behind the main agent's decision to abandon a route.

\textbf{Paper writing.} Danus treats the manuscript as a new mathematical artifact: the completed draft is passed back through the verifier as a whole and revised until it passes as written. Fact-level verification does not transfer, because turning the graph into linear prose creates new mathematics that was never a fact---compressed steps, ``it suffices to'' reductions, glue between facts whose statements nearly but not exactly meet---so correct facts can be stitched into an incorrect manuscript, with the errors at the seams. When a manuscript is too long for a single pass, the main agent decomposes it into self-contained parts, each culminating in a designated result and carrying the results it relies on as established statements, so that the verifier judges a complete document, never a fragment. Where a verified proof was compressed into an assertion, the original is re-supplied and rendered in full. The case of Section~\ref{sec:res-tangent} records this loop in a full run.

\section{Results}\label{sec:results}

To test Danus, we collaborated with a number of mathematicians and conducted a series of experiments in the mode of Danus--human collaboration. Here we summarize a selection of representative papers to illustrate the capabilities of Danus. As a baseline, each of the problems below was also posed, independently of Danus, to the web interface of GPT-5.5-pro; in no case did it produce a meaningful result.

To be precise about the mode of collaboration: these papers were written through interactions in which human experts directed Danus in natural language. In each case, Danus's work ran through the two stages described in Section~\ref{sec:method}: the proof search, which ends when the target theorem stands as a verified fact in the fact graph, and the writing stage, in which the system turns the fact graph into a manuscript (Section~\ref{sec:writing}); human experts could intervene at either stage. Part of the human input was common to every case: the problem statement itself; operational instructions carrying no mathematical content, such as requests for status summaries and the instruction to produce the manuscript; a final check of the completed manuscript, in which human experts confirmed each proof; and revisions that the human authors made independently of Danus after each manuscript was complete, chiefly fine adjustments to the mathematical notation and the citation format (see Section~\ref{sec:limitations}). We call this the common input; the ``Human input'' part of each subsection records only the input that went beyond it.

\subsection{Optimal bend-and-break for foliations}\label{sec:res-bnb}

\textbf{The problem.} Mori's bend-and-break is one of the founding tools of birational geometry: on a smooth projective variety $X$ of dimension $n$ with an ample divisor $H$, through every point of a curve $C$ with $K_X\cdot C<0$ it produces a rational curve of $H$-degree at most $(n+1)\,\frac{H\cdot C}{-K_X\cdot C}$, and the constant $n+1$ is optimal~\cite{Mor82,MM86,JLR25}. For a foliation $\mathcal F\subset T_X$ of rank $r$ (a structure that decomposes the variety, away from a singular locus, into a family of $r$-dimensional leaves), the natural analogue asks for the optimal constant for rational curves \textit{tangent} to $\mathcal F$, that is, curves running along the leaves. Explicit but non-optimal constants were known---$2n$ by Shepherd-Barron~\cite{SB92} and $2r$ by Bogomolov--McQuillan~\cite{BM16}---and the optimal constant remained unknown. The paper~\cite{LiuSunJiang26} settles it: the optimal constant is $r+1$.

\begin{theorem}[{Optimal bend-and-break for foliations~\cite{LiuSunJiang26}}]
Let $X$ be a normal projective variety of dimension $n$, let $\mathcal F$ be a foliation on $X$ of rank $r$, and let $H_1,\dots,H_{n-1},H$ be ample divisors on $X$. Let $C$ be a general complete intersection of elements of $|m_iH_i|$ with $m_i\gg 0$, and suppose $K_{\mathcal F}\cdot C<0$. Then through a general point of $C$ there is a rational curve $\Sigma$ tangent to $\mathcal F$ with
\[
H\cdot\Sigma\ \le\ (r+1)\,\frac{H\cdot C}{-K_{\mathcal F}\cdot C},
\]
and the constant $r+1$ is optimal.
\end{theorem}

The optimal constant depends only on the rank of the foliation, not on the ambient dimension; it matches the classical bound when $r=n$ and the $r=1$ case, i.e. the case of rank $1$ foliations, and once a cone theorem for foliations is available it yields the optimal length bound $r+1$ for $K_{\mathcal F}$-negative extremal rays.

\textbf{How Danus produced it.} A human expert gave Danus the problem and directed it to read two references, the optimal bend-and-break bounds of Jovinelly--Lehmann--Riedl~\cite{JLR25} and the work of Bogomolov--McQuillan~\cite{BM16}. Danus judged on its own that the notation of~\cite{BM16} was not compatible with that of~\cite{JLR25}, and replaced it with the work of Kebekus--Sol\'a Conde--Toma~\cite{KST07}. Danus then completed the proof; five workers explored the problem in parallel, and the search recorded 63 verified facts and 239 failed paths on the way. Danus turned the verified fact graph into a manuscript, and on examining it, a human expert suggested that Danus use the algebraicity criterion of Campana--P\u{a}un~\cite{CP19} to simplify the proof; Danus carried out the revision and finalized the manuscript.

\textbf{Human input.} Beyond the common input, human input consisted of the two initial references and one suggestion arising from the final check of the manuscript: human experts confirmed the correctness of every detail of the proof but found one section more complicated than necessary, and a human expert suggested the simplification noted above; Danus carried out the revision.

\textbf{What this case shows.} In this case the strategy was supplied by the human experts, and Danus's contribution was the execution. First, it carried out a sustained synthesis, transporting an intricate technique from ordinary varieties into the foliated setting on its own, together with the connecting arguments this required. Second, when the references it was given proved unsatisfactory, it sought out a better one on its own.

\subsection{Shokurov's global index conjecture for threefold foliations}\label{sec:res-shokurov}

\textbf{The problem.} Shokurov's global index conjecture predicts that numerically trivial log Calabi--Yau structures are torsion, with index bounded uniformly in the dimension and the coefficient set (see, e.g.,~\cite{Xu20}). In plainer terms, the canonical class of such a structure meets every curve in degree zero, and the conjecture predicts that a single positive integer multiple of it is linearly zero, with the integer depending only on the dimension and the coefficient set. Its foliated version, formulated in~\cite{LMX24} for foliated log Calabi--Yau triples, is known in dimension two~\cite{Per05} and open in dimension three. The paper~\cite{LiuQin26fsi} resolves the three-dimensional case.

\begin{theorem}[{Foliated index theorem in dimension three~\cite{LiuQin26fsi}}]
Let $\Gamma\subset[0,1]\cap\mathbb Q$ be a set satisfying the descending chain condition. Then there is a positive integer $I=I(\Gamma)$ such that for every log canonical foliated triple $(X,\mathcal F,B)$ with $\dim X\le 3$, coefficients of $B$ in $\Gamma$, and $K_{\mathcal F}+B\equiv 0$, one has $I(K_{\mathcal F}+B)\sim 0$. Moreover, if $\mathcal F$ is canonical, not algebraically integrable, and $B=0$, one may take $I\le 30$.
\end{theorem}

This completely resolves the threefold foliated version of Shokurov's global index conjecture. Note that Shokurov's global index conjecture for usual varieties remains open in dimension $\geq 4$, so this is the best foliated result one expects to get now in this direction.

\textbf{How Danus produced it.} A human expert gave Danus the problem. Unprompted, Danus decomposed the problem according to the rank and the algebraic rank of the foliation (the dimension of its leaves, and the dimension of its largest subfoliation with algebraic leaves) into five classes, a classification the human experts considered sound. For three of the five classes, Danus found a Lie-theoretic method and solved them; the approach was one the human experts had not anticipated. It then attempted a similar method on the remaining two classes, the algebraically integrable ones (those whose leaves are algebraic), without success. In the human experts' view, these two classes are precisely the ones for which humans already have a clear and simple approach, while the three classes Danus solved are the harder ones. A human expert then suggested handling the remaining two classes with the foliated minimal model program (the foliated version of birational geometry's standard simplification machinery, recently developed by the human expert and others), and with this hint Danus solved them. The human experts then asked Danus to prove a stronger result---generalizing the index theorem from canonical foliations with $B=0$ to log canonical foliated triples $(X,\mathcal F,B)$ with DCC coefficients---and gave hints toward the ACC for minimal log discrepancies on surfaces~\cite{Ale93} and certain results of Spicer~\cite{Spi20}. Danus completed the generalization. Across the run, seven workers explored in parallel, the first wave running for about eight hours, and the fact graph grew to 784 verified facts, of which 77 form the supporting closure of the theorem. Danus then turned the verified fact graph into the manuscript.

\textbf{Human input.} Beyond the common input, human input consisted of two interventions, both during the proof search: when Danus stalled on the algebraically integrable classes, a human expert suggested the foliated minimal model program, after which Danus resolved them; and the human experts asked for the stronger result, supplying the hints noted above, upon which Danus completed it.

\textbf{What this case shows.} This time the structure of the attack was Danus's own. First, it decomposed the problem into an appropriate classification without any human hint. Second, it discovered a method of solution from a direction the human experts had not expected. Third, when it did stall, a brief hint from a human expert was enough: it understood the suggestion and solved the remaining cases.

\subsection{Total Cartier indices of rational singularities in families}\label{sec:res-cartier}

\textbf{The problem.} The total Cartier index of a normal projective variety is the least common multiple of the Cartier indices of all $\mathbb Q$-Cartier Weil divisors on it---here a divisor is $\mathbb Q$-Cartier if some positive multiple of it is cut out by a single equation near every point, and its Cartier index is the least such multiple. The behavior of this index in families is a basic question about how singularities vary, of the kind the moduli theory of higher-dimensional varieties (in the wake of the KSBA program~\cite{KSB88,Ale94}) constantly needs. Han and Jiang proved boundedness in bounded families of klt-type varieties (a standard class of mild singularities) by running the minimal model program in families, and asked (Problem~4.5 of~\cite{HanJiang26}) whether the much weaker assumption of rational singularities suffices. Rational singularities form a natural frontier here: the class is not stable under the operations of the minimal model program---the tool behind Han and Jiang's result and behind most boundedness theorems of this kind---and without some hypothesis the index can be infinite (a cone over an elliptic curve already fails). The paper~\cite{LiuHuQin26} answers the question affirmatively.

\begin{theorem}[{Total Cartier indices in families~\cite{LiuHuQin26}}]
Let $k$ be an algebraically closed field of characteristic zero and $\pi\colon X\to B$ a projective morphism to a finite-type $k$-scheme. Then there exists a positive integer $I$ such that for every closed point $b\in B$ for which $X_b$ is normal, projective, of pure dimension, and has rational singularities, the total Cartier index of $X_b$ divides $I$.
\end{theorem}

\textbf{How Danus produced it.} A human expert gave Danus the problem, posed at first as a hunt for a counterexample. Seven workers explored refutational and constructive routes in parallel, and the search converged on the affirmative answer: Danus independently discovered a route that synthesizes techniques from markedly different fields: it reduced the algebro-geometric problem to a problem in commutative algebra, and reduced the core of that problem, in turn, to a problem in real algebraic geometry, which it treated with a Hadamard-inequality bound (a classical determinant inequality) among other tools. Along this route Danus also classified the problem, on its own, into the surface (codimension-two) case and the case of dimension at least three (codimension at least three), attacked the two by different methods---controlling the link of the singularity (the boundary of a small neighborhood of the singular point) for surfaces, and using Koll\'ar's work on local Picard groups~\cite{Kol16} in higher dimension---and stitched the two together at the end. Danus then reported that it had completed every case except the codimension three case, which it could not settle. A human expert examined the run and found the cause: unable to obtain the \TeX{} source of~\cite{Kol16}, Danus had worked from the PDF and misread the hypothesis ``dimension $\ge 3$'' as ``dimension $>3$.'' Once this was pointed out, Danus repaired the argument and completed the proof search, and the system turned the fact graph into a manuscript. The first draft was poor: in compressing the verified material into prose, the writing rendered a key lemma in real algebraic geometry incorrectly, while the verified facts beneath it were sound. Submitted back to the verifier (Section~\ref{sec:writing}), the draft was rejected; Danus revised it against the verifier's findings and brought the manuscript to completion.

\textbf{Human input.} Beyond the common input, human input consisted of a single correction: a human expert traced Danus's failure in the three-dimensional case to the misread hypothesis in~\cite{Kol16}, and the repair itself was carried out by Danus.

\textbf{What this case shows.} The origination widened here, in both field and scale. First, Danus found its tools far from the problem's home topic---the minimal model program---and, in the experts' view, changed their sense of the methodology available for such questions. Second, it decomposed the problem on its own and combined the cases at the end. Third, it carried out a long chain of reasoning whose steps are macro-scale reductions, first to commutative algebra and then to real algebraic geometry, rather than lemma-sized ones. Fourth, because the exploration runs refutational and constructive routes together (Section~\ref{sec:workflow}), a search launched toward a counterexample could end by settling the question affirmatively. Fifth, when the writing stage compressed verified material incorrectly, the flaw was caught by verification and repaired without human intervention (Section~\ref{sec:writing}).

\subsection{Factorial asymptotics of the Matryoshka numbers}\label{sec:res-matry}

\textbf{The problem.} The cosmohedron is a positive geometry---informally, a geometric object generalizing a convex polytope, whose shape encodes quantities of physical interest---introduced by Arkani-Hamed, Figueiredo and Vaz\~ao to encode the wavefunction of the universe in a class of cosmological models~\cite{AFV25}. Its combinatorics~\cite{AAFV26} involves the \textit{Matryoshka numbers} (OEIS A177384~\cite{OEISA177384}), defined by $a_1=1$ and $a_n=\sum_{k=1}^{n-1}(k+1)a_ka_{n-k}$. Kot\v{e}\v{s}ovec conjectured, in the OEIS entry, that $a_n\sim c\cdot n!\,n^4$ with $c\approx 0.0054283$; the conjecture is restated as Conjecture~5.7 of~\cite{AAFV26}, whose authors write that they do not know how to prove it. The conjecture asserts that the numbers grow factorially, and pins the growth law down to the exact multiplicative constant. The note~\cite{Liu26matry} proves it.

\begin{theorem}[{Matryoshka asymptotics~\cite{Liu26matry}}]
There is a real number $S$ with $0<S<\infty$ such that
\[
\lim_{n\to\infty}\frac{a_n}{(n+4)!}\;=\;\lim_{n\to\infty}\frac{a_n}{n!\,n^4}\;=\;S,
\qquad
0.00542831750\ \le\ S\ \le\ 0.00542831848 ,
\]
an enclosure of width less than $10^{-9}$ containing the conjectured value.
\end{theorem}

The proof is elementary and every constant in it is explicit, so the enclosure---an interval proved to contain $S$---is rigorous rather than numerical; since $S$ has no known closed form, such an enclosure is the strongest meaningful determination of the constant.

\textbf{How Danus produced it.} A human expert gave Danus the problem. Danus produced the complete proof entirely on its own: it proved that the limit exists and derived explicit upper and lower bounds for it, giving the certified enclosure of $S$ stated above. The run went essentially straight to the proof: five workers, about ninety minutes from launch to the verified target theorem, a fact graph of 100 verified facts with dependency chains up to 15 deep, of which 37 form the supporting closure of the theorem, and a global memory recording a single proof attempt and no dead ends. Danus then turned the verified fact graph into the manuscript, likewise without human involvement.

\textbf{Human input.} Human input did not go beyond the common input: no mathematical guidance of any kind was provided.

\textbf{What this case shows.} Here there was no human mathematical contribution: Danus completed the entire pipeline, from the problem statement to the finished manuscript, without any hint. We note also that the problem comes from the combinatorics of mathematical physics: Danus's range is not confined to a single area of mathematics.

\subsection{Weighted homogeneity via logarithmic vector fields}\label{sec:res-ms}

\textbf{The problem.} A classical theorem of Saito~\cite{Sai71} characterizes weighted homogeneous isolated hypersurface singularities algebraically ($f\in J(f)$, the Jacobian ideal); a hypersurface singularity is weighted homogeneous if, in suitable coordinates, its defining polynomial becomes homogeneous once the coordinates are given appropriate positive weights. da Silva Machado and Seade~\cite{MachadoSeade26} conjectured two characterizations of this hidden symmetry that are geometric rather than algebraic, phrased in terms of vector fields and their behavior near the singular point. A pair of companion papers resolves the conjecture affirmatively: one written entirely by hand~\cite{LiuZhang26human}, the other produced by Danus~\cite{LiuZhang26ai}.

\begin{theorem}[{Weighted homogeneity via logarithmic vector fields~\cite{LiuZhang26human,LiuZhang26ai}}]
Let $(D,0)\subset(\mathbb C^{n+1},0)$ be a reduced isolated hypersurface germ, with $n\ge 2$, or with $n=1$ and $D$ irreducible. Then the following are equivalent: \textup{(i)} $(D,0)$ is weighted homogeneous in suitable coordinates; \textup{(ii)} in suitable coordinates, there is a holomorphic logarithmic vector field everywhere transverse, in the real-Euclidean sense, to all small links of $D$; \textup{(iii)} there is an ambient holomorphic vector field tangent to $D$ with a nondegenerate isolated singularity at the origin.
\end{theorem}

Both new characterizations are geometric, where Saito's criterion is algebraic.

\textbf{How Danus produced it.} A human expert gave Danus the problem, without suggesting any route. Danus completed a proof and turned the verified fact graph into a manuscript. In parallel, a human expert produced an independent proof of the same conjecture; the two proofs proceed by genuinely different methods, and the expert remarked that Danus's approach was one they had not anticipated. On examining Danus's manuscript, the human experts found a gap: the term ``nilpotent'' is used inconsistently across the literature, the definition in a reference Danus had relied on is itself erroneous, and the flaw had propagated into one step of the argument. After a human expert pointed this out, Danus discarded the affected steps, located a correct reference in place of the flawed one, revised the proof substantially, and completed it correctly. The search ran far wider than the proof it finally kept: seven workers, over two waves, produced 687 verified facts---with 23 more revoked in the repair just described---in dependency chains up to 21 deep, while the supporting closure of the main theorem holds about ten facts; the global memory records 91 proof attempts and 49 counterexamples.

\textbf{Human input.} Beyond the common input, human input consisted of one finding in the final check of the manuscript: the human experts identified the flawed reference noted above and pointed it out to Danus. The subsequent overhaul of the proof was carried out by Danus itself.

\textbf{What this case shows.} This case tested Danus against a defect in the literature itself. First, it produced a proof independently, and by a route the human expert had not anticipated. Second, it repaired its own literature dependency: the facts resting on the flawed reference were revoked through the mechanism of Section~\ref{sec:factgraph}, a correct reference was found, and the proof was rebuilt from the verified remainder.

\subsection{Tangent classes of matroids and wonderful compactifications}\label{sec:res-tangent}

\textbf{The problem.} When a matroid $M$ of rank $d+1$ is realized by a complex linear subspace $L$, the wonderful models of De Concini--Procesi~\cite{DCP95} provide smooth projective compactifications $W_{L,\mathcal G}$ of dimension $d$, one for each building set $\mathcal G$ of flats of $M$, and Feichtner--Yuzvinsky~\cite{FY04} presented the Chow ring of $W_{L,\mathcal G}$ by generators $x_F$, indexed by the flats $F\in\mathcal G$, and relations that involve only the matroid $M$. The presentation is intrinsic to the matroid $M$ and therefore defines a graded ring $A_{\mathbb Q}(M,\mathcal G)$, with top-degree map $\deg_{M,\mathcal G}$, for every loopless matroid (not necessarily realizable); its maximal-building-set case is the ring in which Adiprasito--Huh--Katz~\cite{AHK18} proved log-concavity for matroids; Larson--Li--Payne--Proudfoot~\cite{LLPP24} constructed the combinatorial K-ring $K(M,\mathcal G)$, generated by classes $\tau_F$.

One basic piece of the geometry had no counterpart in this dictionary: the tangent bundle of $W_{L,\mathcal G}$. The problem is to construct a tangent class $T_{M,\mathcal G}$ in the integral combinatorial K-ring of every loopless matroid and every building set containing the top flat $E$, such that (i) $T_{M,\mathcal G}$ specializes to the class of the actual tangent bundle of the wonderful compactification $W_{L,\mathcal{G}}$ for a realizable matroid $M$ realized by $L$, (ii) its Hirzebruch--Riemann--Roch numbers recover the graded dimensions of the Chow ring, and (iii) its Chern numbers against the combinatorial hyperplane class are bounded below by $\binom{d+1}{k}$, their value for projective $d$-space. The construction and key properties of such tangent classes were recently studied by Cheng~\cite{Cheng26}. The paper of this case study~\cite{Matroid} constructs the class first in the rational K-ring and proves the three key properties mentioned above, and then adapts it to the integral version, independently of~\cite{Cheng26}, to which, as described below, it had no access.

\begin{theorem}[{Integral tangent classes for matroid wonderful models~\cite{Matroid}}]
Let $M$ be a loopless matroid of rank $d+1$ on a finite nonempty ground set $E$, and let $\mathcal G$ be a Feichtner--Yuzvinsky building set on $M$ containing the top flat $E$. Then there is a class $Q_{M,\mathcal G}^{\mathbb Z}$ in the integral combinatorial K-ring $K_{\mathbb Z}(M,\mathcal G)$ such that
\[
T_{M,\mathcal G}^{\mathbb Z}:=\sum_{F\in\mathcal G\setminus\{E\}}(1-\tau_F)^{-1}-Q_{M,\mathcal G}^{\mathbb Z}
\]
is a genuine element of $K_{\mathbb Z}(M,\mathcal G)$, with integer coordinates in the standard $\tau$-monomial basis, and, writing $T_{M,\mathcal G}$ for its rationalization: \textup{(i)} if $M$ is realized by a complex linear subspace $L$, then $T_{M,\mathcal G}^{\mathbb Z}$ maps to the class of the tangent bundle of $W_{L,\mathcal G}$ under an integral ring isomorphism $K_{\mathbb Z}(M,\mathcal G)\xrightarrow{\sim}K_0(W_{L,\mathcal G})$; \textup{(ii)} for every $0\le i\le d$,
\[
\dim_{\mathbb Q}A_{\mathbb Q}(M,\mathcal G)^i=(-1)^i\deg_{M,\mathcal G}\!\left(\operatorname{ch}\bigl(\wedge^i T_{M,\mathcal G}^{\vee}\bigr)\operatorname{td}\bigl(T_{M,\mathcal G}\bigr)\right),
\]
with $\operatorname{ch}$ and $\operatorname{td}$ taken in the formal splitting-root sense; \textup{(iii)} for every $0\le k\le d$,
\[
\deg_{M,\mathcal G}\!\left(c_k(T_{M,\mathcal G})\,\alpha^{d-k}\right)\ \ge\ \binom{d+1}{k},
\qquad \alpha:=-x_E .
\]
\end{theorem}

\textbf{How Danus produced it.} The problem was posed, in a fixed verbatim prompt and with no suggested route, to three systems: the web interface of GPT-5.5-pro, Rethlas, and Danus. The experiment was run before~\cite{Cheng26} appeared on arXiv, and that paper was deliberately withheld from all three, so the problem they faced was open. Neither of the first two produced a solution. Rethlas, run on the problem three times, returned nothing that passed the verifier; its attempts foundered precisely on the conflations the problem invites, for instance between a realizable matroid, its wonderful model and the permutohedral toric variety, or between integral and rational K-theory. Danus, which runs the same workers against the same verifier, solved the rational form of the problem and then adapted it to the integral version. Seven workers explored constructive and refutational routes in parallel for about five days. The search was far broader than the proof it left behind: the final fact graph, shown in Figure~\ref{fig:factgraph}, holds 3,157 verified facts in dependency chains up to 54 facts deep, of which 664 form the supporting closure of the theorem, and the global memory records 636 proof attempts, 151 counterexamples, and 25 dead ends. Danus then turned the verified fact graph into a manuscript; reorganizing the verified material into linear prose introduced errors of its own, so the draft was passed back through the verifier and corrected section by section. After the human experts observed that the completed manuscript solved only the rational version of the problem, whose original statement had asked for an integral class, Danus resumed work and produced a solution to the integral version. Details of the experiment are recorded in \cite[Appendix B]{Matroid}.

\textbf{Human input.} Beyond the common input, human input consisted of a single mathematical intervention, which came after the system had declared the task finished: the human experts pointed out that the solution addressed the rational rather than the integral version of the problem, upon which Danus produced the integral version. The human authors then checked the manuscript and found it correct, with one local exception: the justification given for \cite[Lemma 8.7]{Matroid} (used in the Chern--\(\alpha\) lower bound) is incomplete as written, the lemma having been treated as proved without a valid argument. The lemma is nonetheless true and admits a short proof (see \cite[Proposition~4.19]{Cheng26}), and this small gap affects neither the construction of the tangent class nor the $P^K=\mathrm{Hilb}$ identity.

\textbf{What this case shows.} The final case is the largest of the six searches and the closest our experiments come to a controlled comparison between Danus and its predecessor: Rethlas and Danus ran the same worker and verifier models on the same prompt, so the difference in outcome, three failed runs against a complete verified solution, reflects the orchestration around the models rather than the models themselves. The comparison isolates two capabilities. First, Danus accommodates a much greater depth of reasoning: the proof grew fact by fact into a verified closure of hundreds of statements, and the long-horizon design let the run continue until the theorem was verified rather than for a preset number of rounds. Second, Danus is sensitive to mathematics whose difficulty lies in fine distinctions: where the single-line attempts of Rethlas conflated nearby notions, in Danus several workers attacked the problem simultaneously from different directions, including attempts at refutation, and every claim had to pass the verifier in the exact form in which later facts would use it, so the distinctions the problem turns on stayed intact across hundreds of steps. The class Danus constructed recovers the tangent class and its three key properties of~\cite{Cheng26}, which it had never seen.

\section{Discussion}\label{sec:discussion}

The case studies of Section~\ref{sec:results} show what the design of Section~\ref{sec:method} delivers: proofs built from hundreds of verified facts, produced in days, with human intervention concentrated at a few decisive points. They show its boundaries just as plainly: the manuscripts still need expert review, and the hardest steps of several searches were unlocked by a human. This section takes up four questions these observations raise: Section~\ref{sec:disc-models} asks why each model holds its role and what the assembly gains over its parts; Section~\ref{sec:disc-fluency}, how the writing stage balances fluency against faithfulness; Section~\ref{sec:disc-verification}, why reliable verification underlies everything else; and Section~\ref{sec:disc-scaling}, in what sense the design scales test-time computation and where the scaling stops. Section~\ref{sec:str_limit} then collects the strengths the experiments exhibited and the limitations they exposed.

\subsection{Comparison with GPT-5.5, Claude Opus 4.8, GPT-5.5-pro, and Rethlas}\label{sec:disc-models}

The assignment of models to roles follows their measured strengths. In our experiments GPT-5.5 showed stronger mathematics than Claude Opus 4.8, so the workers and the verifier run on Codex agents with GPT-5.5. Claude Code with Claude Opus 4.8, in turn, was better suited to the main agent's task of reading a large, evolving body of logs, memory, and a fact graph of thousands of entries (Section~\ref{sec:mainagent}). GPT-5.5-pro has the strongest mathematics of the three but is too costly to call continuously, so it serves as the low-frequency expert reference. We have not yet systematically evaluated the more recent Fable 5 from Anthropic.

Assembled this way, the system is stronger than any of its parts. That the harness can matter more than the raw model was already visible in Rethlas: an agent running on GPT-5.4 and GPT-5.5 solved a research-level problem in algebraic groups on which GPT-5.5-pro, through its web interface, produced a completely wrong proof (Section~5.1.1 of~\cite{Ju+26}). Moreover, no single model call is designed to return a verified proof of the length our case studies require. Even GPT-5.5-pro offers at best an outline of a direction, often an erroneous one; when posed our problems directly, it produced no meaningful result in any case (Section~\ref{sec:results}).

Much of the design can therefore be read as machinery for eliciting from each model more than a direct call obtains. The verifier holds the reasoning to mathematical rigor: context isolation and skills that spell out the checking logic largely suppress hallucination and skipped steps, which no direct call achieves, even one explicitly instructed to check its result before answering (Section~\ref{sec:workers}). Repeated calls and independent workers absorb the errors that even the strongest models make: a GPT-5.5-pro consultation enters the memory as guidance, never the fact graph as truth; a wrong direction is tried, refuted, and recorded as a dead end, while a right one, even one in ten, is carried through into verified facts. The search thus profits from GPT-5.5-pro's strength at whatever reliability it offers. Parallelization scales the search with test-time computation, and the fact graph and the memory extend its length: each agent handles a small task while the system advances on the large one (Section~\ref{sec:disc-scaling}).

The assembled system thus outperforms not only its models called alone but also Rethlas itself, which continues inside Danus essentially as a single worker--verifier loop: on the same problem, with the same models, Rethlas failed three times where Danus produced the verified solution (Section~\ref{sec:res-tangent}).

\subsection{Balance between writing fluency and truth faithfulness}\label{sec:disc-fluency}

Faithfulness to the proof record and readability pull in opposite directions, and the tension grows with the size of the fact graph. The most faithful manuscript would simply write out, one after another, every fact the target theorem depends on; it would be correct, since each fact has passed verification, and unreadable, running to hundreds of facts at the scale of our runs. A readable paper instead needs motivation, an order of presentation different from the order of discovery, and the omission of routine detail. But this rewriting is itself mathematical writing, and it introduces errors at a rate far above the graph's own. The characteristic failure, seen repeatedly in our experiments, is a manuscript whose results, constructions, and route are all correct while the written reasoning goes wrong exactly where several facts were compressed into one (Sections~\ref{sec:writing} and~\ref{sec:res-tangent}).

The remedy is to pursue the two goals in sequence: the draft is written for readability, and the completed manuscript must then pass the verifier as written. Until it does, it is revised and resubmitted, the worker's submit--verify--repair cycle lifted to the whole paper, except that here the main agent faces the verifier and writes no new facts: each revision retrieves existing facts from the graph and renders them more accurately (Sections~\ref{sec:workers} and~\ref{sec:writing}). This removes most writing-stage errors while the paper remains a paper, not a fact-by-fact transcript.

What remains is a balance rather than a resolution: the manuscripts can still explain motivation too thinly or compress steps too far, and demanding more rigor pushes the text back toward the transcript. The released versions are the point we chose, and the residual defects are small: the human revisions recorded in Section~\ref{sec:results} were chiefly adjustments of notation and citation format.

\subsection{Importance of verification}\label{sec:disc-verification}

Danus takes its verifier from Rethlas and deliberately narrows its role: an independent, stateless service concerned with nothing but the correctness of the text in front of it (Section~\ref{sec:workers}). The isolation is essential: prompting a model to check its own work suppresses hallucination only mildly; the checker must be a separate agent whose context holds the submission, the cited facts, and nothing else, and whose skills prescribe the checking procedure. Engineered this way, the verifier almost never accepts a flawed proof: across our experiments the few slips on record were caught in the final human review (Section~\ref{sec:limitations}), and revocation, the graph's repair channel, was needed only once (Section~\ref{sec:res-ms}).

This precision is what makes the fact graph workable. A worker trusts every existing fact without re-deriving it (Section~\ref{sec:factgraph}), so a fact is only as reliable as everything beneath it. The theorem of Section~\ref{sec:res-tangent} rests on 664 supporting facts in chains up to 54 deep, and a structure of this depth stands or falls with the verifier: if verification is imprecise, errors accumulate along the chains, the facts higher up become unreliable, and long-horizon reasoning is impossible; if it is precise, accumulation is safe. The verifier's precision is thus the load-bearing assumption of the architecture.

Verification also changes the economics of the human--machine loop. In the writing loop of Section~\ref{sec:disc-fluency} it acts as a first referee: what it rejects is repaired by the machine and never reaches the expert. This matters because human checking, not machine search, is the bottleneck; results produced in a day or two typically took experts a week or two to check. Verification, more than search speed, is what accelerates the joint production of new mathematics.

\subsection{Test-time scaling in width and in depth}\label{sec:disc-scaling}

The design began with a simple intention: Rethlas spends its test-time computation on one line of reasoning at a time, and we wanted to spend more of it, on many lines at once. Simply adding workers fails: all of them edit the single blueprint carrying the whole proof, and each one's progress is every other's interference (Section~\ref{sec:factgraph}). The fact graph makes the parallel work add up: it splits the proof into facts a single worker can own, so contributions accumulate instead of colliding, and the main agent spreads the workers over different lines of attack (Sections~\ref{sec:factgraph} and~\ref{sec:mainagent}). In the run of Section~\ref{sec:res-tangent}, seven workers built 3,157 verified facts in five days; only 664 of them support the theorem, and the rest record how widely the system searched. This is scaling in width: added computation becomes added exploration.

Width alone buys many short arguments rather than one long one; length comes from trusting what is already built and remembering what has already been tried. Verification provides the trust: every fact is checked on admission (Section~\ref{sec:disc-verification}), so a worker can stand on a chain 54 facts deep without re-deriving any of it. Memory does the remembering: raw exploration stays in the local logs, distilled lessons rise into the global memory, and the main agent's summaries compress the state of the run into a page (Sections~\ref{sec:memory} and~\ref{sec:mainagent}). Each level condenses the one below it, so what any agent reads stays bounded while what the system knows grows: a worker on its tenth task inherits the lessons of the first nine from a summary, not five days of logs. This is scaling in depth: an effective reasoning horizon far beyond a single context window.

The machinery behind both axes is modest, a main agent, a fact graph, and the interaction logic between them, yet it separates the case studies of Section~\ref{sec:results} from what the same models, or Rethlas itself, produce alone (Section~\ref{sec:disc-models}). The same experiments also mark where the scaling stops. The scaling works where a path to the solution exists and needs only to be found: width locates the entry points, depth carries the argument through. It does not create the path: when the solution demands an idea beyond what any single call proposes, more workers and longer horizons merely let the system circle, accumulating shallow conclusions that never break the problem open. In our experiments such impasses were broken by a human supplying the missing idea (Section~\ref{sec:res-shokurov}); for the deepest open problems there may be no one to supply it, and attacking them will need agent frameworks more inventive than the present one.

\subsection{Strengths and limitations}\label{sec:str_limit}\label{sec:limitations}

The preceding subsections each pursued a single question; to close the discussion, we summarize Danus's capabilities as a whole. Danus showed many strengths in our experiments. First, the quality of the verifier was satisfactory: we observed only a handful of verifier errors in total, most of which stemmed from imprecise or erroneous statements in the cited references (Section~\ref{sec:res-ms}). The fact graph could therefore serve reliably as the system's source of truth. On this foundation, with the main agent directing the search and the memory carrying its lessons forward, Danus extended its work in depth and in width: it carried an intricate technique into a new setting in one sustained synthesis (Section~\ref{sec:res-bnb}), kept a long parallel search running until the target theorem stood verified (Section~\ref{sec:res-tangent}), and decomposed problems unprompted into parts worked along constructive and refutational routes (Sections~\ref{sec:res-shokurov} and~\ref{sec:res-cartier}). In work of this breadth and length, Danus showed notable mathematical capability: at times it completed an entire proof with little or no human assistance, in one case the whole pipeline from problem statement to finished manuscript (Section~\ref{sec:res-matry}); it reached solutions by routes the human experts had not anticipated (Sections~\ref{sec:res-shokurov} and~\ref{sec:res-ms}), drew tools from branches of mathematics distant from the problem's own (Section~\ref{sec:res-cartier}), and, when a supplied reference was suboptimal, sought out a better one on its own (Section~\ref{sec:res-bnb}). Finally, Danus displayed a measure of writing ability: when a first draft rendered a key lemma incorrectly, the verifier rejected the draft and Danus repaired it automatically (Section~\ref{sec:res-cartier}); another manuscript was corrected section by section the same way (Section~\ref{sec:res-tangent}).

At the same time, the experiments exposed several shortcomings. There remain cases in which Danus does not find the correct route to a proof on its own, and a hint from a human expert is then needed (Section~\ref{sec:res-shokurov}). Even when the route is found, Danus's paper writing is not yet fully satisfactory: citation formatting occasionally strays from the conventions of the field, and its notation, though logically consistent, does not always guard against confusion; these imperfections required manual adjustment by the human experts afterwards. In searching the literature, Danus tends to settle for references that are mathematically sufficient, rather than the ones that would make a proof simplest: the simplifying criterion in the case of Section~\ref{sec:res-bnb} had to be pointed out by a human expert. Finally, Danus relies on the information it extracts from its references, and can inherit an error when a reference is itself flawed; errors of this kind are fully resolved only by expert review: in the case of Section~\ref{sec:res-ms}, an erroneous definition propagated into the argument until the experts caught it and the facts resting on it were revoked (Section~\ref{sec:factgraph}).

\section{Conclusion}\label{sec:concl}
In this paper, we propose Danus, an orchestration system for research-level mathematical reasoning centered on a shared fact graph as a global memory-management mechanism. Danus separates global planning, parallel proof search, and verification: a main agent coordinates the overall strategy, multiple worker agents explore and prove local claims in parallel, and a stateless verifier checks proposed claims before they are admitted into the fact graph. By storing each verified fact together with its proof and logical dependencies, the fact graph allows Danus to accumulate long mathematical arguments incrementally while keeping the shared proof state organized and reliable.

We evaluated Danus through six research-level case studies in algebraic geometry, singularity theory, and combinatorics. These examples show that Danus can construct long and detailed proofs, decompose difficult problems into workable subproblems, repair flawed proof dependencies, and turn verified fact graphs into mathematical manuscripts. The amount of human input varied across the case studies, ranging from no mathematical guidance to a small number of high-level hints or corrections, but in all cases Danus carried out substantial proof construction.

Our results suggest that fact-graph-based orchestration provides an effective mechanism for extending LLM-based mathematical reasoning agents to long-horizon research problems. By turning parallel local proof search into an organized and cumulative proof-building process, Danus enables many contributions from different workers to be assembled into coherent mathematical arguments. At the same time, our experiments show that human mathematicians remain essential for choosing meaningful problems, identifying missing ideas, and carrying out final expert review. We therefore view Danus as a step toward collaborative mathematical research systems in which AI agents can contribute substantially to proof construction while remaining integrated with human mathematical judgment.

\section*{Acknowledgements}

This work is supported in part by the Fundamental and Interdisciplinary Disciplines Breakthrough Plan of
the Ministry of Education of China (JYB2025XDXM113), the National Key R\&D Program of China grant
2024YFA1014000, and the New Cornerstone Investigator Program.

\bibliographystyle{plainnat}
{\hbadness=1100
\bibliography{ref}}

\end{document}